\documentclass[10pt,conference,a4paper]{IEEEtran}
\usepackage{url}
\usepackage{subfig}
\usepackage{float}
\usepackage{graphicx}
\usepackage{amsmath}
\usepackage{amssymb}
\usepackage{bm}

\begin{document}
\title{Efficient Text Classification Using Tree-structured Multi-linear 
Principal Component Analysis}

\author{\IEEEauthorblockN{Yuanhang Su, Yuzhong Huang and C.-C. Jay Kuo}
\IEEEauthorblockA{University of Southern California, Los Angeles, California, USA\\
Email: \{yuanhans,yuzhongh\}@usc.edu cckuo@sipi.usc.edu}}

\maketitle

\begin{abstract}
A novel text data dimension reduction technique, called the
tree-structured multi-linear principal component analysis (TMPCA), is
proposed in this work. Being different from traditional text dimension
reduction methods that deal with the word-level representation, the
TMPCA technique reduces the dimension of input sequences and sentences
to simplify the following text classification tasks. It is shown
mathematically and experimentally that the TMPCA tool demands much lower
complexity (and, hence, less computing power) than the ordinary
principal component analysis (PCA).  Furthermore, it is demonstrated by
experimental results that the support vector machine (SVM) method
applied to the TMPCA-processed data achieves commensurable or better
performance than the state-of-the-art recurrent neural network (RNN)
approach. 
\end{abstract}

\IEEEpeerreviewmaketitle
\section{Introduction}\label{sec:intro}

Text classification has been an active research topic for about two
decades.  Its implementations such as spam email detection, age/gender
identification and sentiment analysis are omnipresent in our daily
lives. Tasks on smaller datasets are generally regarded as simple ones,
which can be typically handled by linear models such as the naive Bayes
\cite{NB} classifier nowadays.  Although text classification is an
intensively studied topic, it still faces challenges as the tasks become
more complex due to ever-increasing data amount and diversity in the
Internet. 

The increased complexity partly comes from diversified text patterns as
a result of a larger vocabulary set and/or more sentence variations of
similar meanings. Such a phenomenon is known as ``the curse of
dimensionality" \cite{Sparsity}. To address the high data dimension
problem, one way to reduce the data dimension lies in an efficient
numericalization (or embedding) of text data.  Typically, dimension
reduction is conducted at the word level so that it is called the word
embedding process. As the accumulated volume of language data throughout
the Internet becomes higher, simple models trained on limited datasets
with word embedding cannot keep up with the complexity of existing tasks
\cite{Conv_text}. 

In this research, we go one step further by embedding the entire input
sequence and/or sentence into one vector while keeping the sequential
patterns as intact as possible with an objective to facilitate the
classification task that follows.  However, the complexity of the whole
sequence/sentence embedding is extremely high. This is only made
possible by introducing some novel technique.  The main contribution of
this work is the proposal of a novel technique, called the
tree-structured multi-linear principal component analysis (TMPCA), to
reduce the dimension of input data efficiently. The TMPCA tool is
applied to whole sequence/sentence embedding so that they can be
effectively trained and tested using machine learning models and tools.
We will show that the TMPCA tool can retain word correlations (i.e.,
text patterns) in the input sentence in a compact form. Furthermore, it
demands much less computing power than the ordinary PCA in the training
process. 

The rest of this paper is organized as follows. Related previous work is
reviewed in Sec.  \ref{sec:related}. Then, the new TMPCA text dimension
reduction technique is presented in Sec.  \ref{sec:TMPCA}. Experimental
results are given in Sec. \ref{sec:exp}, where we compare the
performance of methods using the TMPCA technique and other
state-of-the-art methods.  Finally, concluding remarks are drawn in Sec.
\ref{sec:conclusion}. 

\begin{figure}[t]
\centering
\includegraphics[width=\linewidth]{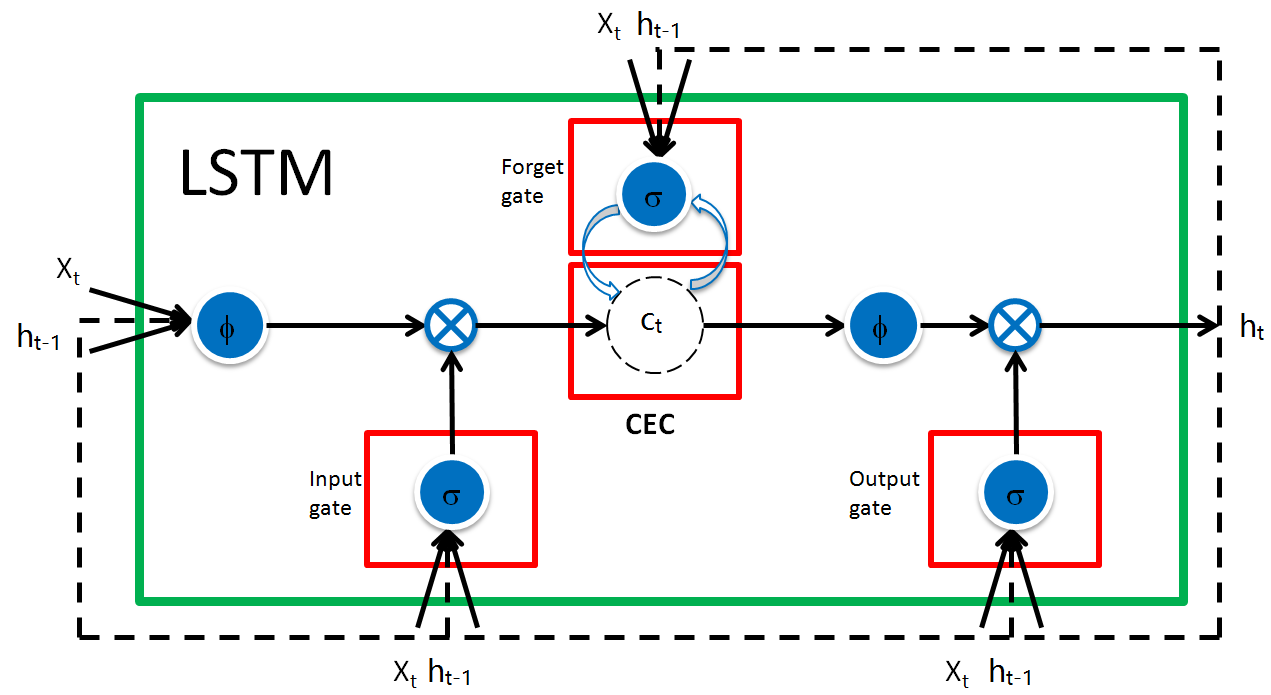} 
\caption{The block diagram of a long short-term memory (LSTM) system.}\label{fig:LSTM}
\end{figure}

\section{Related Previous Work}\label{sec:related}

The idea of representing text patterns (or numericalized sentences) with
a compact vector is explored in the development of recurrent neural
networks (RNN). Such a representation is stored as a hidden state of
RNN's basic computing unit known as the memory cell. There are two
popular cell designs: the long short-term memory (LSTM) \cite{LSTM} and
the gate recurrent unit (GRU) \cite{GRU}.  The block diagram of an LSTM
system is shown in Fig. \ref{fig:LSTM}.  As shown in this figure, each
cell takes the element from a sequence as its input sequentially and
computes an intermediate value that can be recurrently updated. Such a
value is called the constant error carousal (CEC) in the LSTM and simply
a hidden state in the GRU.  Usually, many cells are connected to form a
complete RNN, and the intermediate value from each cell forms a vector
called the hidden state. 

It was observed in \cite{Time} that, if a hidden state is properly
trained, it can represent the desired text patterns compactly and group
similar semantic word level features closely.  This property was
further analyzed in \cite{DBRNN-ELSTM}. Generally speaking, for a well
designed representational vector (namely, the hidden state), the
computing system (i.e., the memory cell) is powerful in exploiting the
word-level dependency to facilitate the final classification task. 

\begin{figure}[t]
\centering
\includegraphics[width=0.7\linewidth]{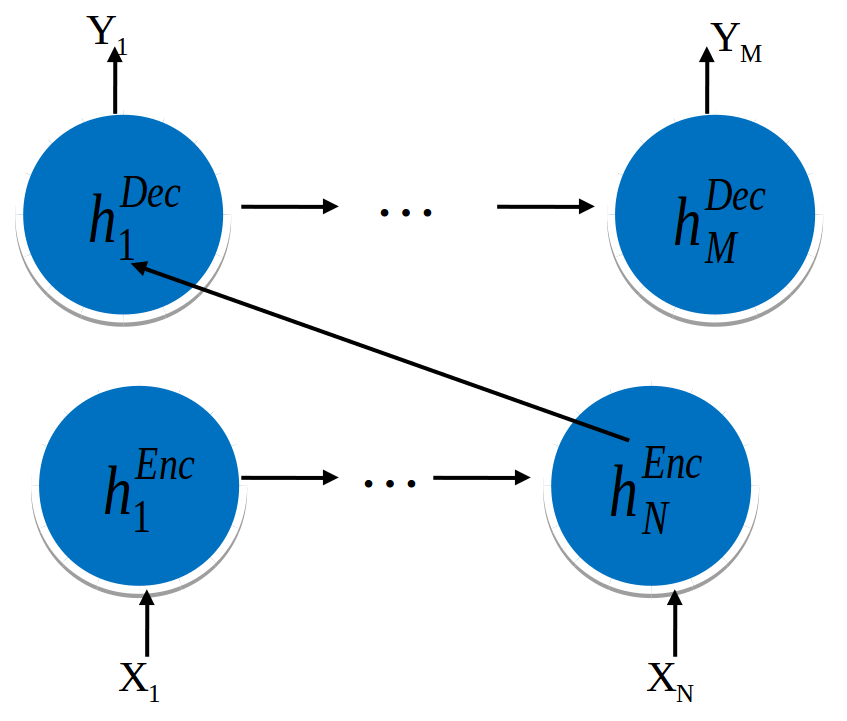} 
\caption{The block diagram of a seq2seq system.}\label{fig:enc-dec}
\end{figure}

The representation power of a hidden state has been utilized in
sequence-to-sequence (seq2seq) learning \cite{Seq2Seq,Grammar}, which is
a variant of the RNN model. Its block diagram is shown in Fig.
\ref{fig:enc-dec}.  The Seq2seq system reduces a higher dimensional
input sequence into a lower dimensional hidden state using an
encoder-decoder structure as shown in Fig.  \ref{fig:enc-dec}. Both the
encoder and the decoder are implemented as RNN models.  The encoder cell
takes an input sequence of variable length and stores the text patterns
in the hidden state. This is known as ``encoding". Given the hidden
state information, the decoder learns how to correlate the input
automatically and generate the desired output in the decoding process. 

Ideas in reducing the numericalized text dimension are also present in
non-neural-network based methods. Traditional text classification often
uses the stop-word list to remove non-informative words such as
prepositions, conjunctions, etc.  The principal component analysis (PCA)
was applied to matrices in \cite{LSA,Noise_reduction}, where each entry
indicates the frequency of a term occurring in a document.  Text
dimension reduction was also conducted in \cite{Two-stage} by removing
words of low information gain and, then, applying the PCA to the
word-level feature space. Words of similar distributions were clustered
and then a naive Bayes classifier was trained for text classification in
\cite{Clustering}. All the aforementioned methods ignore the word
positional information and adopt the ``bag-of-words" representation in
dimension reduction. Apparently, the sequential correlation of words in
sentences is lost in such a treatment. 

\section{Tree-structured Multi-linear PCA (TMPCA)}\label{sec:TMPCA}

In this section, we propose a new technique for text data dimension
reduction and name it ``tree-structured multi-linear principal component
analysis (TMPCA)".  As compared with traditional text dimension
reduction methods that focus on the word-level representation, the TMPCA
technique is designed to reduce the dimension of the entire sentences or
sequences while preserving the sequential order of composing words. It
generates a more compact representation than the hidden state of RNNs.
By eliminating the redundant information in sentences/sequences, it
alleviates the overfitting problem in classifier training. We will
elaborate the detailed design below. 

\begin{figure*}[htb]
\centering
\includegraphics[width=0.9\linewidth]{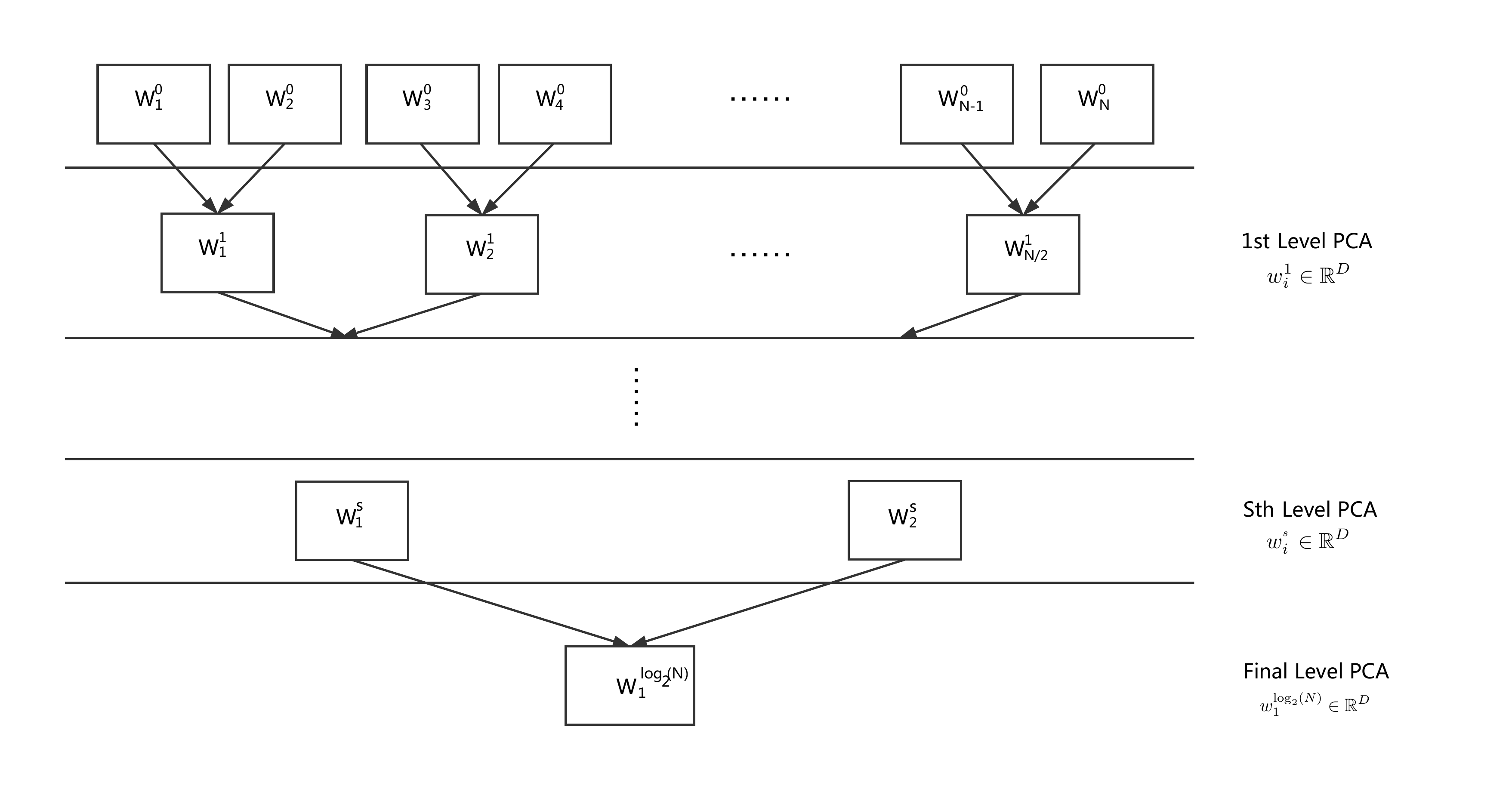} 
\caption{The block diagram of TMPCA}\label{fig:TMPCA}
\end{figure*}

\subsection{Proposed TMPCA Algorithm}

The TMPCA method reduces the sentence length yet keeps the word
embedding size. Its block diagram is shown in Fig. \ref{fig:TMPCA}.
Suppose that each element at the top level, denoted by $W_n^0$,
$n=1,\cdots,N$, has an embedding dimension of $D$. At each level of the
tree, every two adjacent elements are concatenated to form one vector of
dimension $2D$ to train the PCA kernel. Multiple input vectors serve as
the rows of the input matrix to the PCA. Each PCA transform two input
vectors from the upper level to its corresponding lower level. The
dimension of the output vector is halved from $2D$ to $D$ by the PCA.
For example, consider a sentence composed by four elements: $\{w^0_1$,
$w^0_2$, $w^0_3$, $w^0_4\}$, where $w^0_i \in \mathbb{R}^D$, $\forall
i\in \{1,2,3,4\}$.  The superscript denotes the level in the TMPCA tree
and level $0$ means the original numericalized sentence/word sequence.
The following data matrix is used to represent the input:
$$
\begin{bmatrix} 
(w^0_1)^T& (w^0_2)^T \\ 
(w^0_3)^T& (w^0_4)^T 
\end{bmatrix}.
$$ 
The output of the first level MPCA can be written as
$$
\begin{bmatrix}
(w^1_1)^T \\ (w^1_2)^T \end{bmatrix}, \quad
w^1_1 = U(\begin{bmatrix} w^0_1 \\ w^0_2\end{bmatrix}), \quad
w^1_2 = U(\begin{bmatrix}w^0_3 \\ w^0_4\end{bmatrix}),
$$ 
and where $U$ is the MPCA's transform matrix defined on the sentence
length dimension, and $U\in \mathbb{R}^{D\times2D}$. The new transformed
sentence at level 1 can be expressed as $\{w^1_1, w^1_2\}$. Then, it
serves as the input to the next level TMPCA transform. It is apparent
that, after one-level TMPCA, the sentence length is halved while the
word embedding size, $D$, keeps the same; namely, $w^1_i \in
\mathbb{R}^D, \forall i \in \{1,2\}$. Such a process is repeated until
the whole sentence is reduced to a single-word vector.  

\begin{table}[htb]
\caption{Change of dimension}\label{tab:dim_change}
\begin{center}
\begin{tabular}{l | c | c}
\hline\rule{0pt}{12pt}
& Sentence length   & Embedding size \\[2pt] \hline\rule{0pt}{12pt}
Input sentence & N & D\\ \hline\rule{0pt}{12pt}
TMPCA Transformed sentence& 1& D \\[2pt] \hline
\end{tabular}
\end{center}
\end{table}

The dimension evolution from the initial input data to the ultimate
transformed data is summarized in Table \ref{tab:dim_change}. The
original input sentence has $N$ words and each word is embedded with a
$D$-dimensional vector.  After the $\log_2(N)$-level TMPCA transform,
the sentence length becomes one word of embedding dimension $D$. 

\subsection{Computational Complexity Analysis}

To analyze the computational complexity of the TMPCA algorithm, we
consider a sentence of length $N=2^L$. The total number of training
sentences is $M$, and the word embedding size is $D$. To fit the PCA model
to this training matrix of dimension $\mathbb{R}^{M \times ND}$ requires
$\text{O}(MN^2D^2)$ to compute the covariance matrix of the dataset, and
$\text{O}(N^3D^3)$ to compute its eigenvalues. 

At level $s$, the dimension of the training matrix is equal to
$M\frac{N}{2^s}\times2D$. Thus, the total computational complexity of
the TMPCA algorithm can be derived as
\begin{align}
\text{O}(f_{\text{TMPCA}}) & = \text{O}\Big(\sum^{\log_2N}_{s=1}\big(( 2D)^3 + 
                                M\frac{N}{2^s}(2D)^2 \big)\Big)\\
			& = \text{O}\Big(8 L D^3 + 4M(N-1)D^2\Big) \\
			& = \text{O}\Big(2 L D^3 + MND^2\Big) \label{eq:TMPCA_complx}.
\end{align} 
The complexity of the traditional PCA can be written as
\begin{align}\label{eq:PCA_complx}
\text{O}(f_{\text{PCA}}) &= \text{O}\Big(N^3 D^3 + M N^2 D^2 \Big).
\end{align} 
By comparing (\ref{eq:TMPCA_complx}) and (\ref{eq:PCA_complx}), we
see that the time complexity of the TMPCA algorithm grows at most
linearly with sentence length $N$. Furthermore, if $N \ll D$,
$\text{O}(f_{\text{TMPCA}})$ grows logarithmly with $N$.  In contrast,
the traditional PCA grows at least quadratically with $N$. Thus,
$f_\text{PCA}$ grows much faster than $f_\text{TMPCA}$, or
$f_{\text{TMPCA}}=\text{o}(f_{\text{PCA}})$

If we concatenate $P$ non-overlapping elements at each 
tree-level, the time complexity is then:
\begin{equation}\label{eq:TMPCA_N_complx}
\text{O}(f_{\text{TMPCA}}) = \text{O}\Big((P^3\log_PN)D^3 + 
MPND^2\Big).
\end{equation}
As shown in Eq. (\ref{eq:TMPCA_N_complx}), the time complexity increases
with $P \in \{2, \cdots , N \}$. The worst case is $P=N$, which is the
traditional PCA applied to the entire sentence. 

One reason to combine two non-overlapping elements at each tree-level is
its computational efficiency. Another is that it can preserve the
sentence structure well. We set all sentences to be of the same length
$N$.  Any sentence of length shorter than $N$ is padded by one or more
special symbols to length $N$.  Sentences of length longer than $N$ will
be truncated to $N$. The sentences are tokenized, and each token
corresponds to an embedded/numercalized word vector. The embedding can
be done by either one-hot vector embedding or word2vec embedding as
reviewed in Sec. \ref{sec:exp}. 

\section{Experimental Results}\label{sec:exp}

\subsection{Experimental Setup}

We conducted experiments on the following four datasets.
\begin{enumerate}
\item SMS Spam dataset (SMS SPAM). It has ``Spam" and ``Ham" as two
target classes. 
\item Standford Sentiment Treebank (SST). It has ``positive" and
``negative" as two target classes. The labels are generated using the
Stanford CoreNLP toolkit \cite{SST_NLP_tool}. The sentences labeled as
very negative or negative are grouped into one negative class. Sentences
labeled as very positive or positive are grouped into one positive
class.  We keep only the positive and negative sentences for training
and testing. 
\item Semantic evaluation 2013 (SEMEVAL). We focus on Sentiment task-A
with positive/negative two target classes. Sentences labeled as neutral
are removed. 
\item Cornell Movie review (IMDB). It contains a collection of movie review
documents with their sentiment polarity - positive or negative. 
\end{enumerate}
More details of these four datasets are given in Table \ref{tab:data}.

\begin{table*}[htb]
\centering
\caption{\textbf{Dataset}}\label{tab:data}
\begin{tabular*}{\textwidth}{@{\extracolsep{\fill} }l | c c c c c}
\hline
& Target Classes & Train Sentences & Development Sentences & Test Sentences & \# 
of Tokens  \\ \hline
SMS SPAM & 2 & 5,574 & 500 & 558 & 14,657 \\ \hline
SST & 2 & 8409 & 500 & 1803 & 18,519\\\hline
SEMEVAL & 2 & 5098 & 915 & 2034 & 25,167 \\\hline
IMDB & 2 & 10162 & 500 & 500 & 20,892 \\\hline
\end{tabular*}
\end{table*}

\begin{table*}[htb]
\centering
\caption{\textbf{Sentence Length}}\label{tab:length}
\begin{tabular*}{\textwidth}{@{\extracolsep{\fill} }l | c c c c }
\hline
                    & SMS SPAM & SST & SEMEVAL & IMDB \\ \hline
Sentence Length (N) & 64 & 64 & 32 & 64 \\ \hline
\end{tabular*}
\end{table*}

The sentence length is fixed for each dataset for the four benchmarking
methods; namely, TMPCA with SVM, PCA with SVM, SVM only and RNN.  The sentence length values are
shown in Table \ref{tab:length}.  To numericalize the text data, we
remove stop words from sentences according to the stop-word list,
tokenize sentences and, then, stem tokens using the python natural
language toolkit (NLTK).  Afterwards, we use the Wiki2vec embedding
\cite{Wiki2vec} to embed stemmed tokens into vectors. The embedding size
is $1000$.  We used the SVM as the classifier and applied it on these
embedded vectors in their raw forms or processed by TMPCA and PCA,
respectively. 

We compare the performance of the following four methods:
\begin{itemize}
\item SVM: Raw embedded features followed by SVM;
\item TMPCA+SVM: TMPCA-processed features followed by SVM;
\item PCA+SVM: PCA-processed features followed by SVM;
\item RNN.
\end{itemize}
The input to these methods is a single long vector by concatenating all
embedded word vectors from a sentence in order. The first three were
trained on Intel Core i7-5930K CPU while the RNN was trained on the
GeForce GTX TITAN X GPU. The setup of the RNN is given in Table
\ref{tab:GRU_setup}. 

\begin{table}[htb]
\caption{RNN setup details.}\label{tab:GRU_setup}
\begin{center}
\begin{tabular}{l | c }
\hline\rule{0pt}{12pt}
RNN model & seq2seq with attention \cite{Grammar}\\[2pt]
\hline\rule{0pt}{12pt}
Cell & LSTM \cite{LSTM} \\ \hline\rule{0pt}{12pt}
Number of layers   & 1 \\ \hline\rule{0pt}{12pt}
Embedding size & 512\\ \hline\rule{0pt}{12pt}
Number of cell & 512\\\hline\rule{0pt}{12pt}
Training steps & 10 epochs \\\hline\rule{0pt}{12pt}
Learning rate & 0.5 \\\hline\rule{0pt}{12pt}
Training optimizer & AdaGrad \cite{AdaGrad} 
\\[2pt] \hline
\end{tabular}
\end{center}
\end{table} 

\begin{table*}[htb]
\centering
\caption{\textbf{Error Rate ($\%$)}}\label{tab:results}
\begin{tabular*}{\textwidth}{@{\extracolsep{\fill} }l | c c c c}
\hline
& TMPCA+SVM & PCA+SVM & SVM & RNN \\ \hline
SMS SPAM & \textbf{2.33} & 2.87 & 2.69 & 2.51 \\ \hline
SST &  \textbf{21.24} & 27.18 & 27.07 & 24.57\\\hline
SEMEVAL & \textbf{24.09} & 24.49 &24.78& 25.17 \\\hline
IMDB & \textbf{24.40} & 25.60 & 29.20& 30.20 \\\hline
\end{tabular*}
\end{table*}

\subsection{Experimental Results}

The error rates of four benchmarking methods for four datasets are shown
in Table \ref{tab:results}, where the best figures are highlighted in
boldface.  We see from Table \ref{tab:results} that, although the data
dimension is reduced to approximately a 32th or 64th of the original
size, the performance of the TMPCA+SVM method does not degrade as much
as that of the PCA+SVM method. This substantiates the claim that TMPCA
is better in preserving the structure of input sentences.  In addition,
the fact that SVM performs better on reduced datasets demonstrates that
the TMPCA is able to remove the weakly or non-correlated information
from the dataset while preserving the principal ones. This helps
alleviate the overfitting problem. 

\begin{figure*}[htb]
\centering
\subfloat[]{\label{fig:time_total_imdb} \includegraphics[width = 
0.35\linewidth]{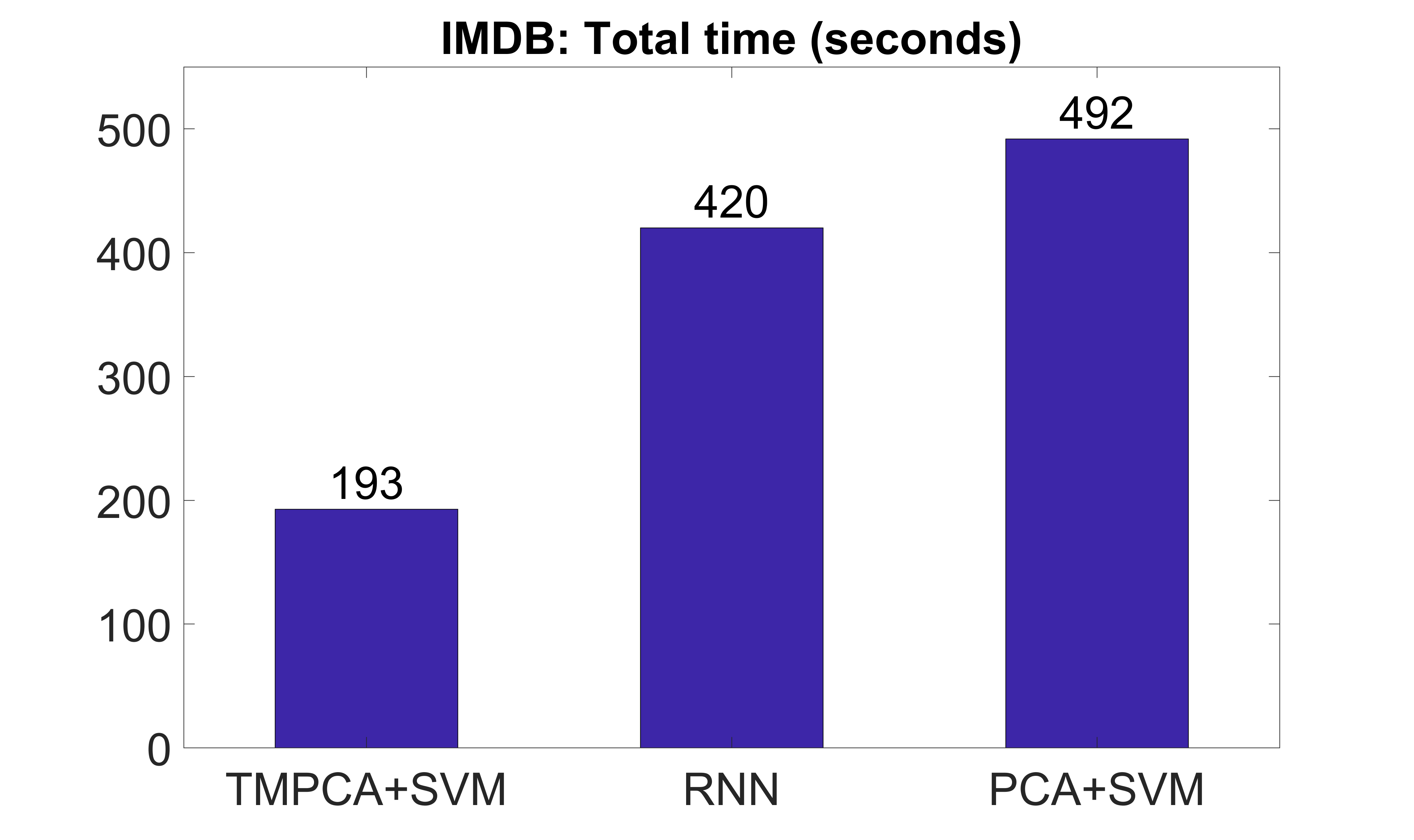}}
\centering
\subfloat[]{\label{fig:time_total_semeval} \includegraphics[width = 
0.35\linewidth]{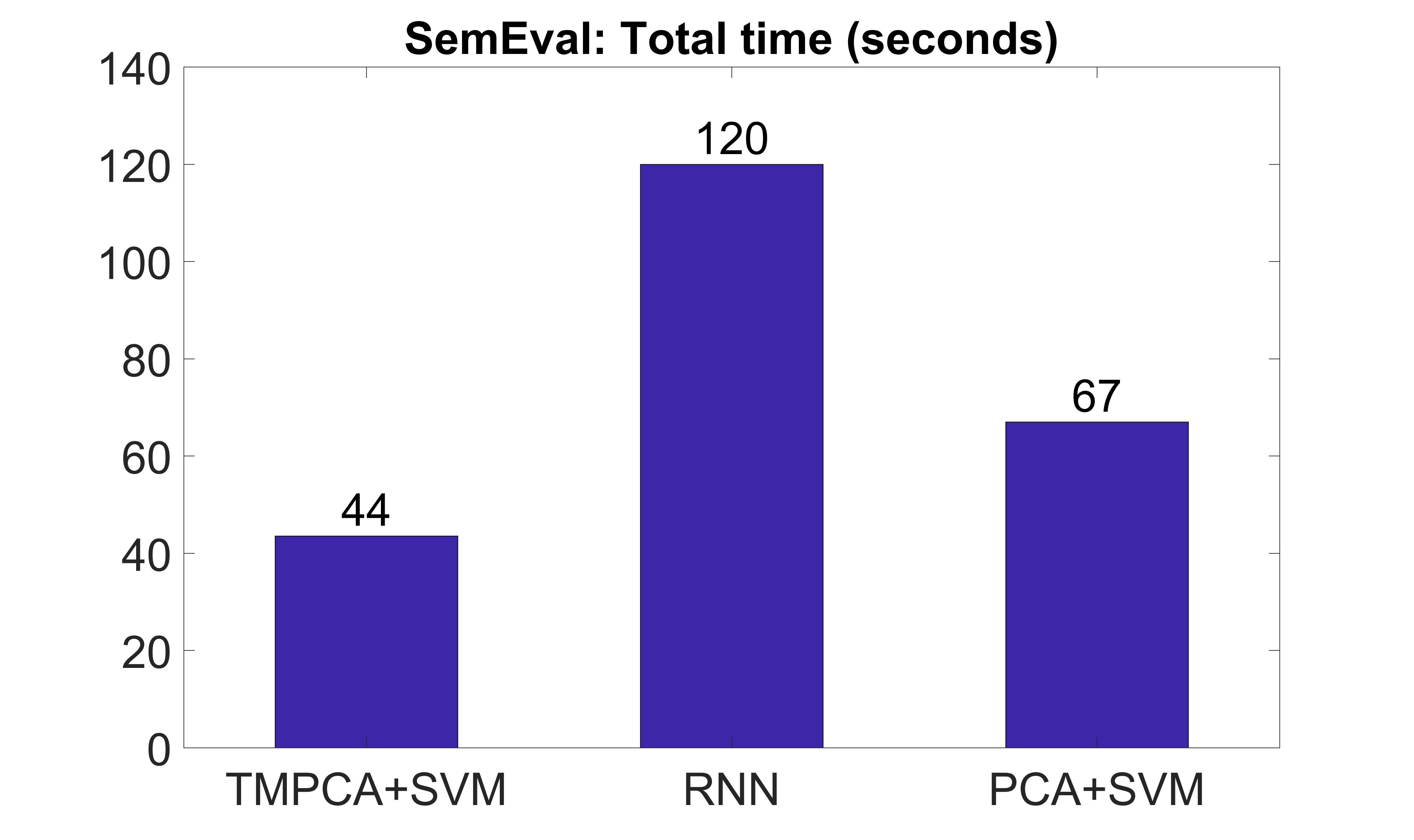}}\\
\centering
\subfloat[]{\label{fig:time_total_sst} \includegraphics[width = 
0.35\linewidth]{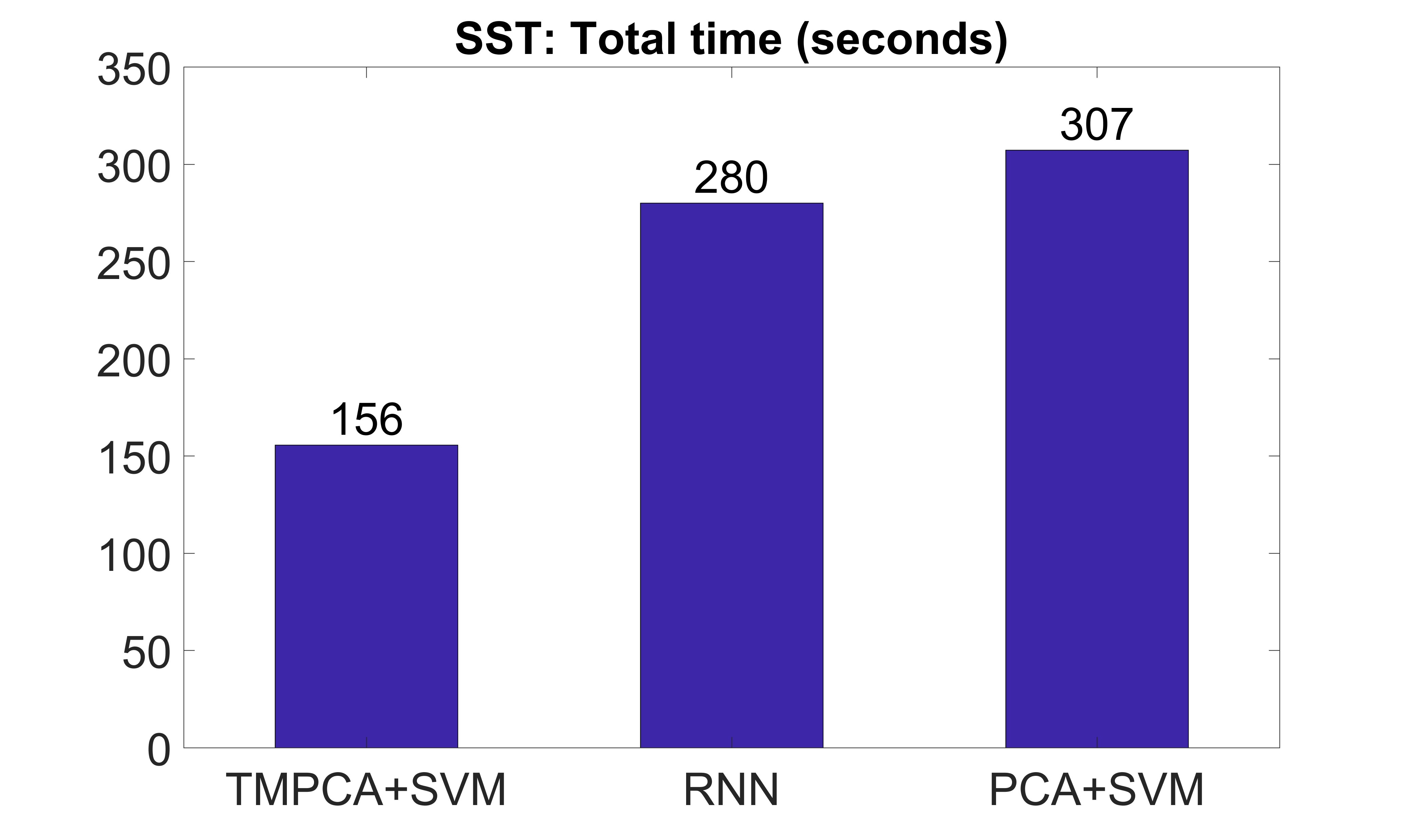}}
\centering
\subfloat[]{\label{fig:time_total_spam} \includegraphics[width = 
0.35\linewidth]{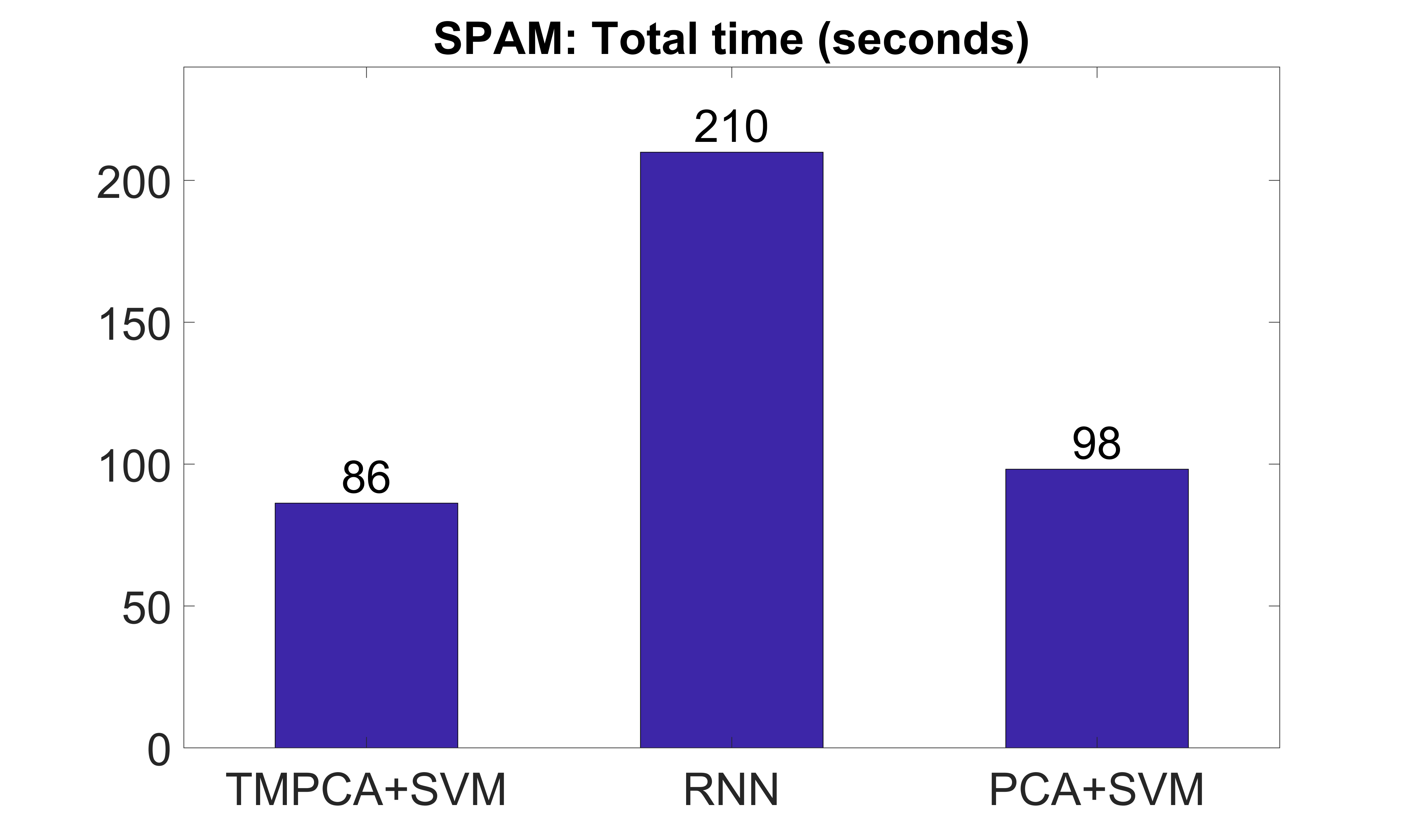}}
\caption{Comparison of the total training time of three methods 
against four datasets.} \label{fig:total_time}
\end{figure*}

We compare the total training time for the TMPCA+SVM, PCA+SVM and RNN
methods against the four datasets in Fig. \ref{fig:total_time}(a)-(d).
Clearly, the total training time taken by TMPCA+SVM is shorter than
PCA+SVM and RNN.  It is also worthwhile to mention that the computation
of TMPCA and SVM was done on the CPU while the RNN was run on the GPU.
The latter is known to be more efficient in large scale data
computation.  

\begin{figure*}[htb]
\centering
\subfloat[]{\label{fig:time_svm_imdb} \includegraphics[width = 
0.35\linewidth]{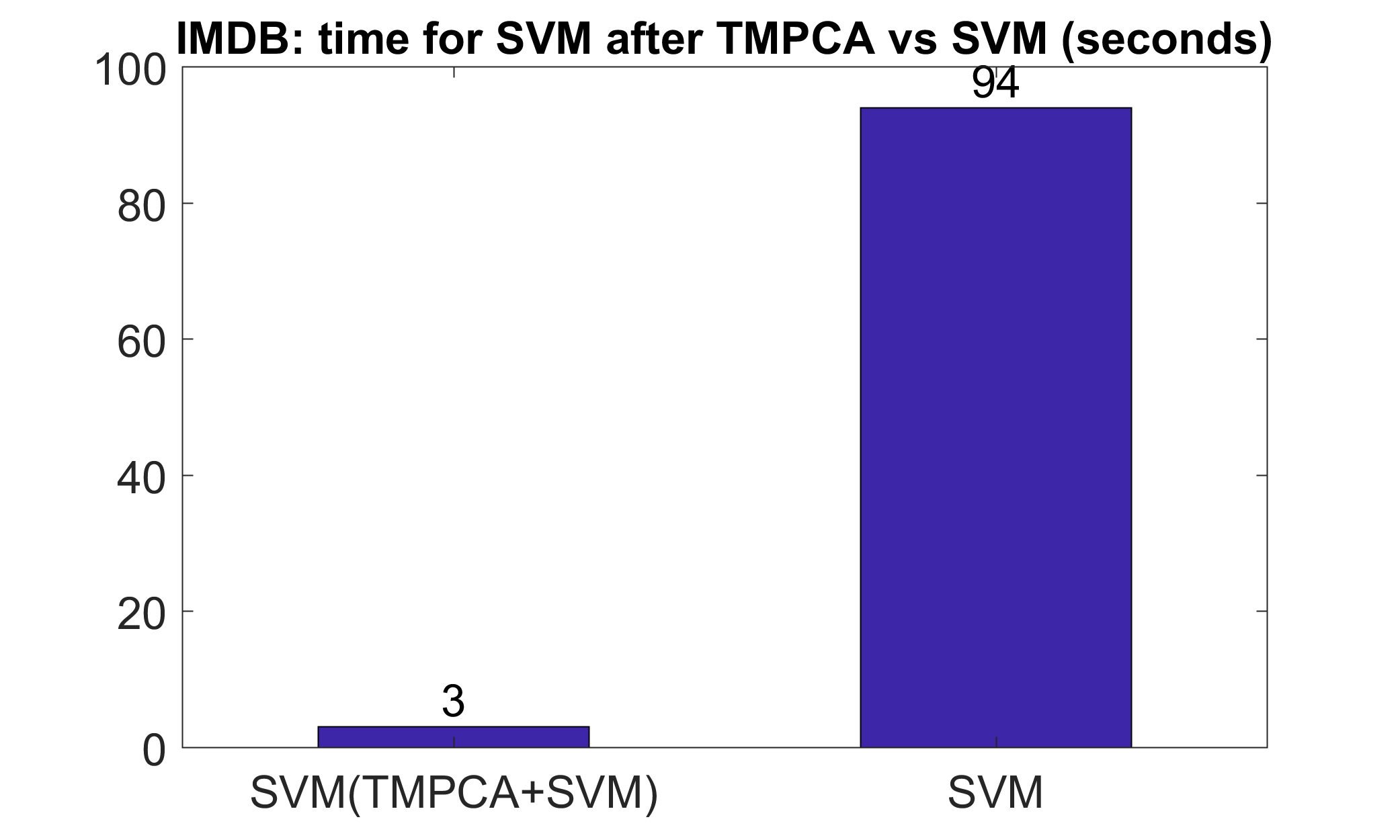}}
\centering
\subfloat[]{\label{fig:time_svm_semeval} \includegraphics[width = 
0.35\linewidth]{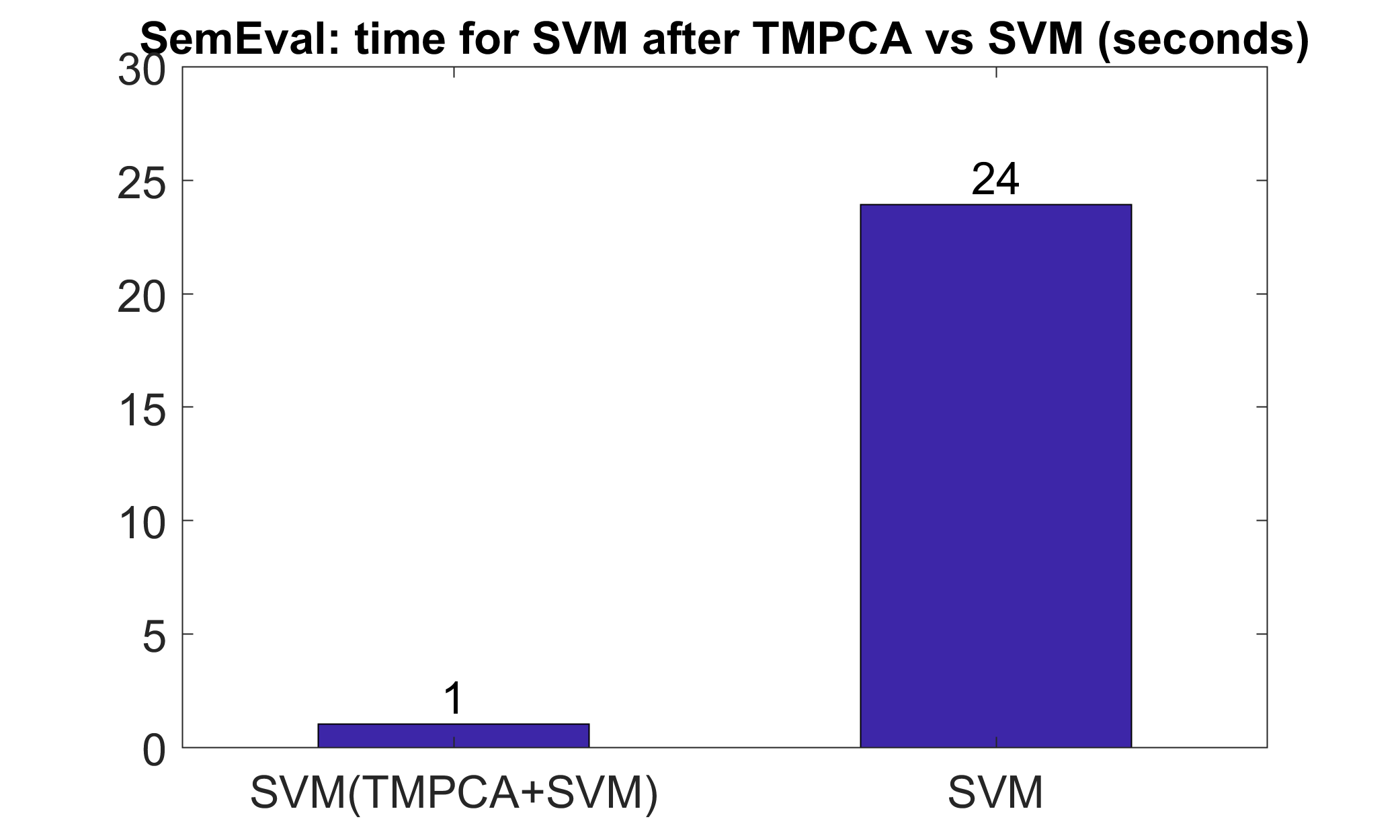}}\\
\centering
\subfloat[]{\label{fig:time_svm_sst} \includegraphics[width = 
0.35\linewidth]{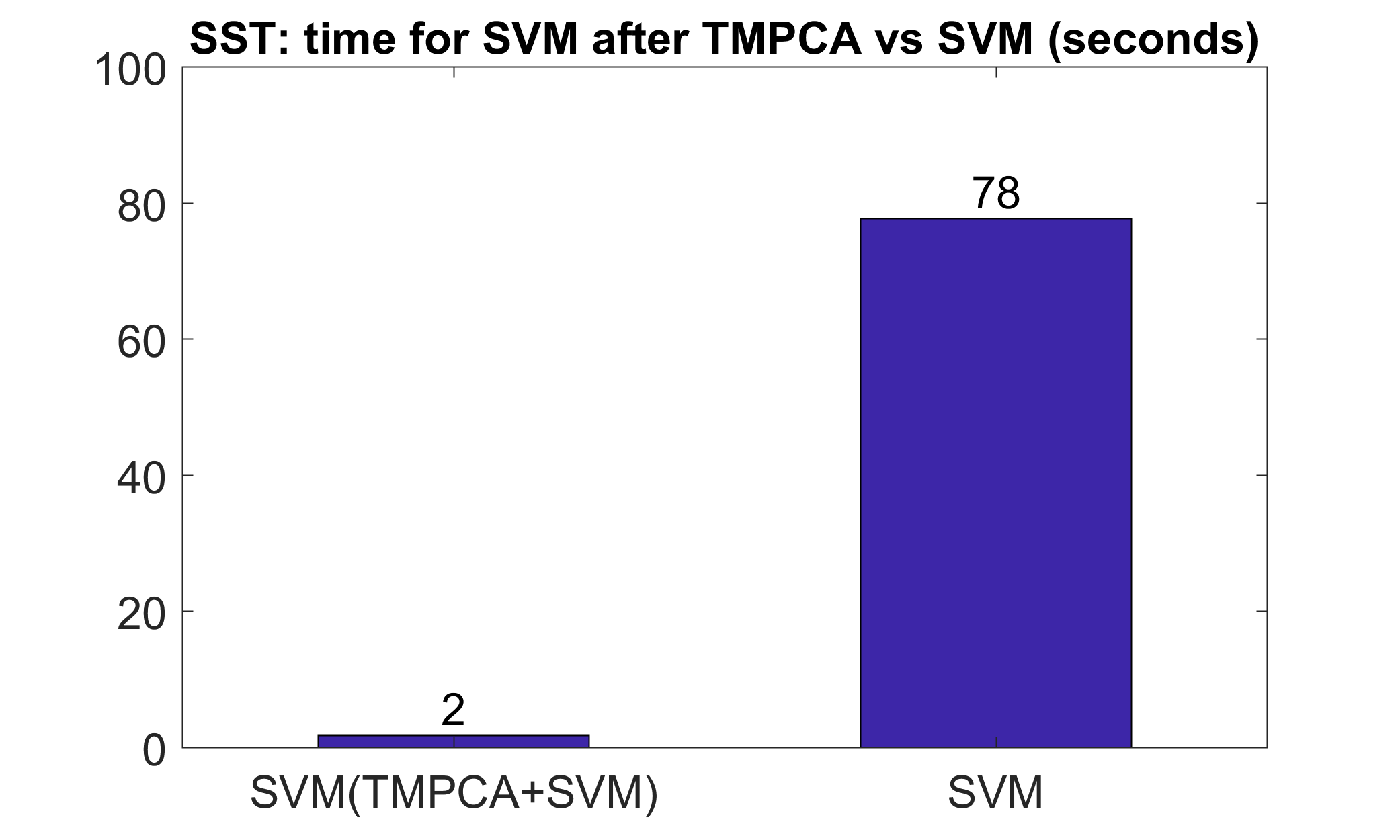}}
\centering
\subfloat[]{\label{fig:time_svm_spam} \includegraphics[width = 
0.35\linewidth]{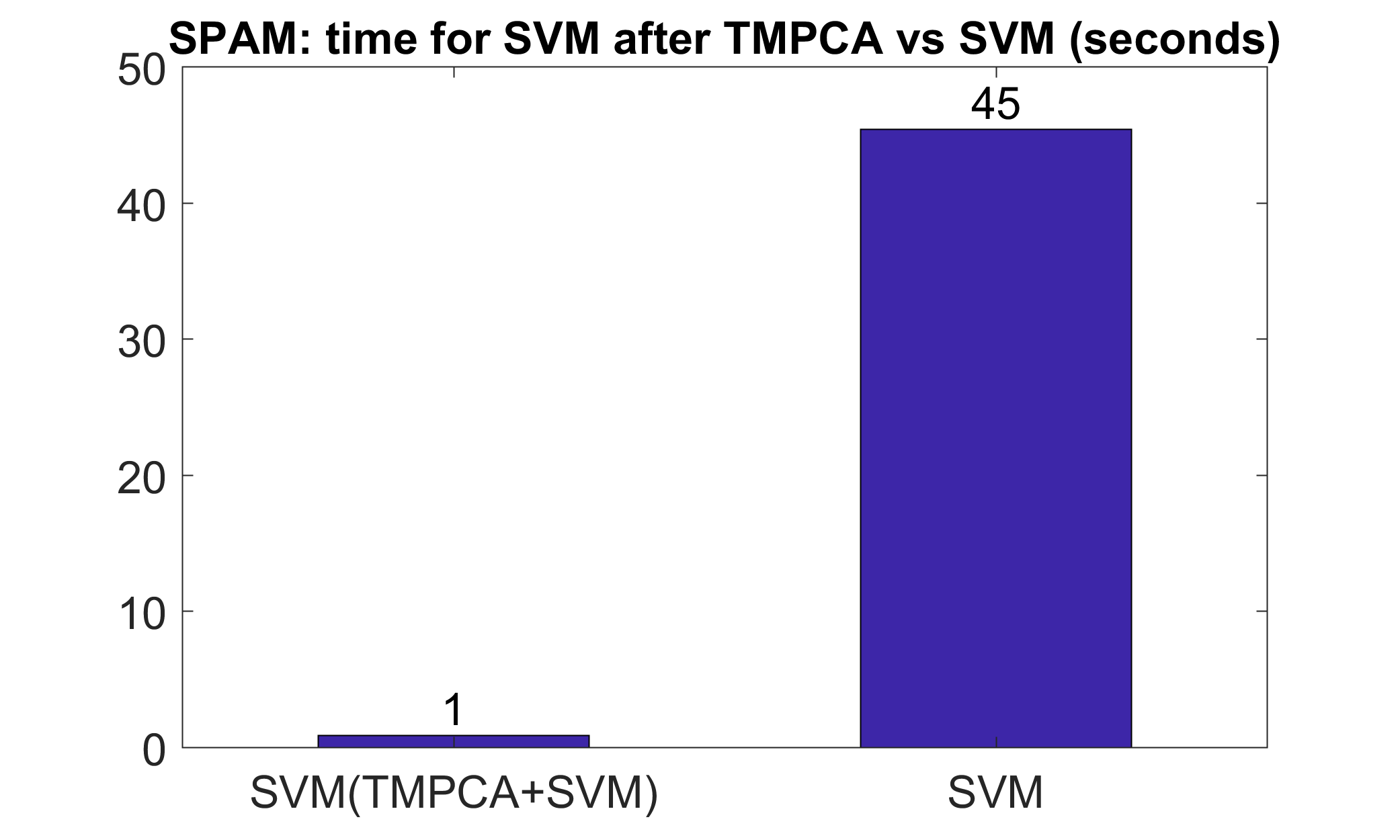}}
\caption{Comparison of the SVM training time based on the TMPCA
processed data and the raw input data.} \label{fig:svm_time}
\end{figure*}

As a result of the reduced dimension, the SVM training time on the TMPCA
processed data is only a fraction of time used by the SVM on the
original data. Their SVM training time is compared in Fig.
\ref{fig:svm_time} (a)-(d). We can see the advantage of data reduction
clearly in training time saving. 

Since the N-gram \cite{Ngrams} technique is popular in text
classification, we would like to see whether it can bring any benefit to
the proposed TMPAC+SVM method. We preprocessed sentences into different
gram forms. For example, the bigram form of a sentence consisting of 4
words denoted by ``1, 2, 3, 4" is ``12, 23, 34". Its 3-gram is in form
of ``123, 234", etc.  Then, we train the TMPCA+SVM model on the N-gram
preprocessed sentences, and test the model on the original test
sentences, which are not N-gram preprocessed, to check the robustness of
the TMPCA method.  The value of $N$ chosen in our experiments were 1,
2, 4 and 8. 

\begin{figure*}[htb]
\centering
\subfloat[]{\label{fig:imdb_ngrams} \includegraphics[width = 
0.35\linewidth]{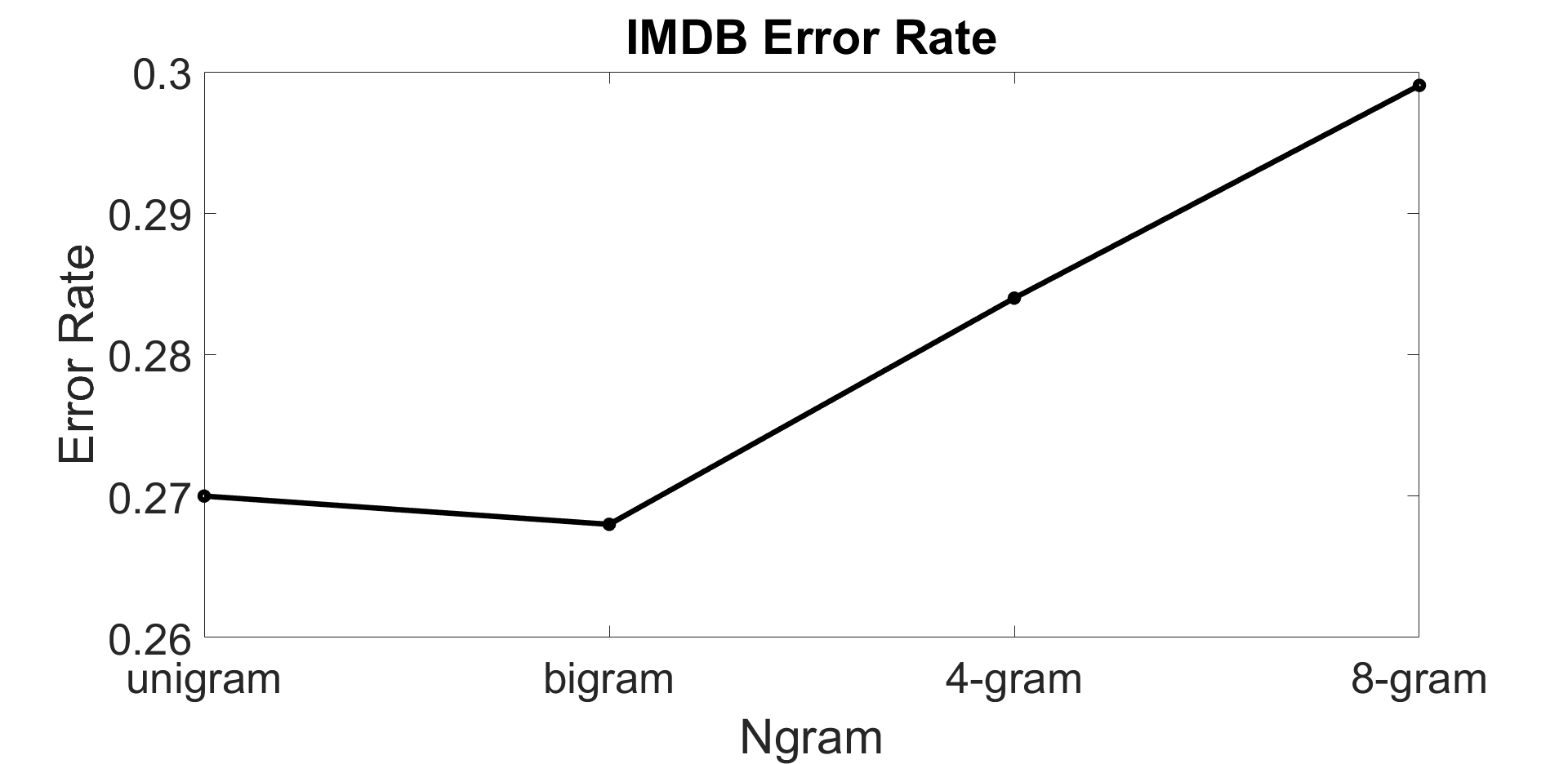}}
\centering
\subfloat[]{\label{fig:semeval_ngrams} \includegraphics[width = 
0.35\linewidth]{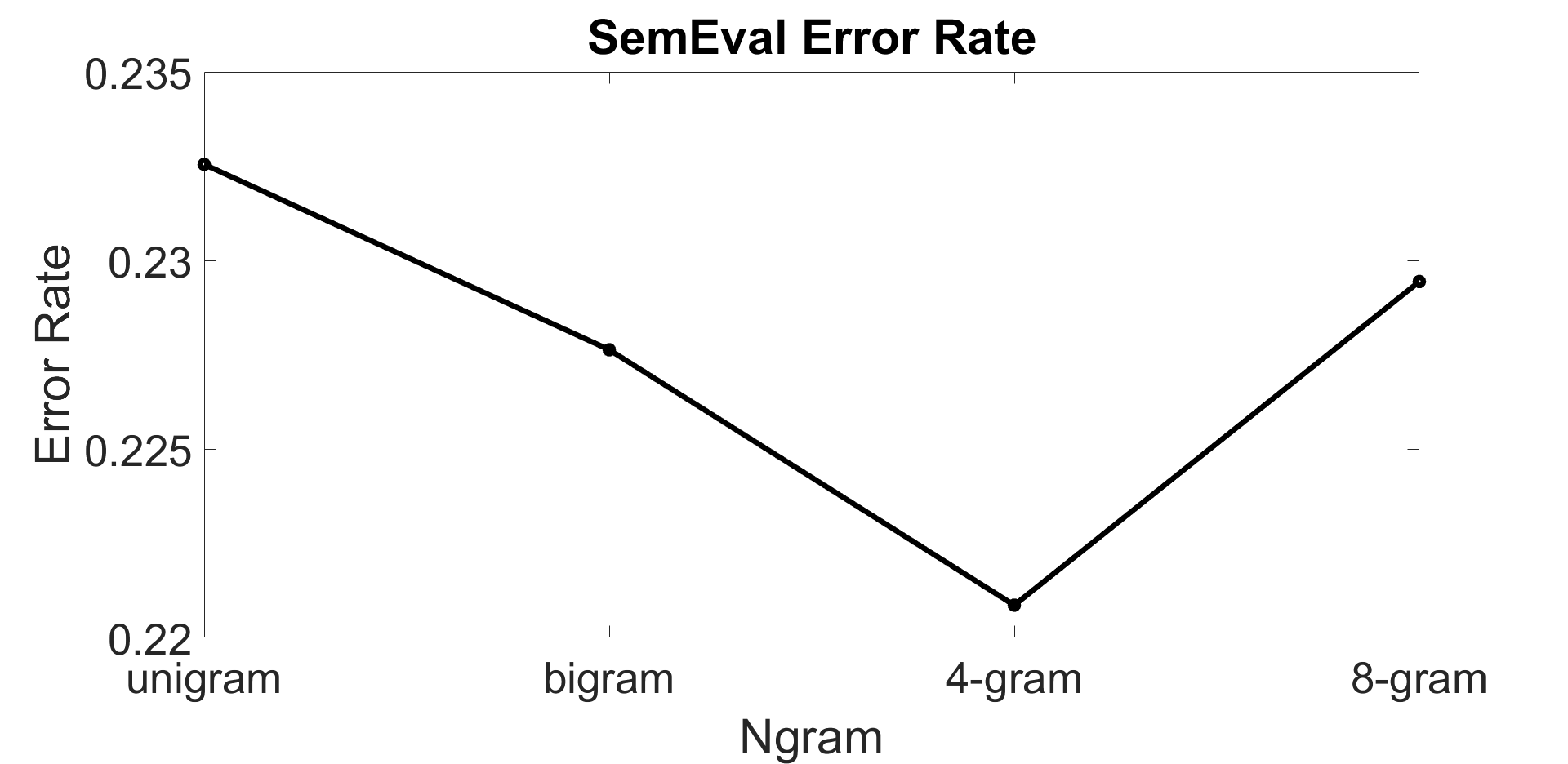}}\\
\centering
\subfloat[]{\label{fig:sst_ngrams} \includegraphics[width = 
0.35\linewidth]{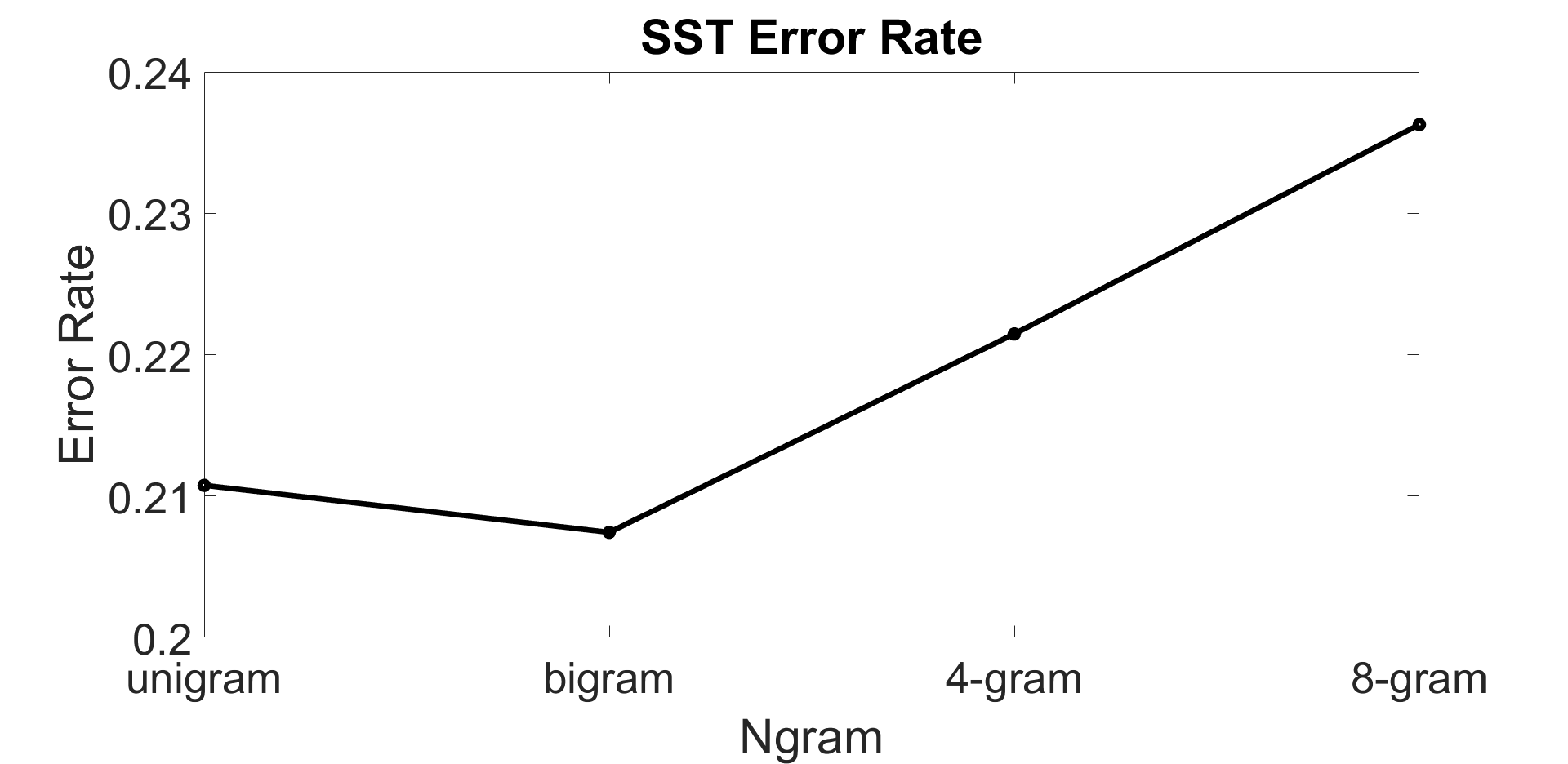}}
\centering
\subfloat[]{\label{fig:spam_ngrams} \includegraphics[width = 
0.35\linewidth]{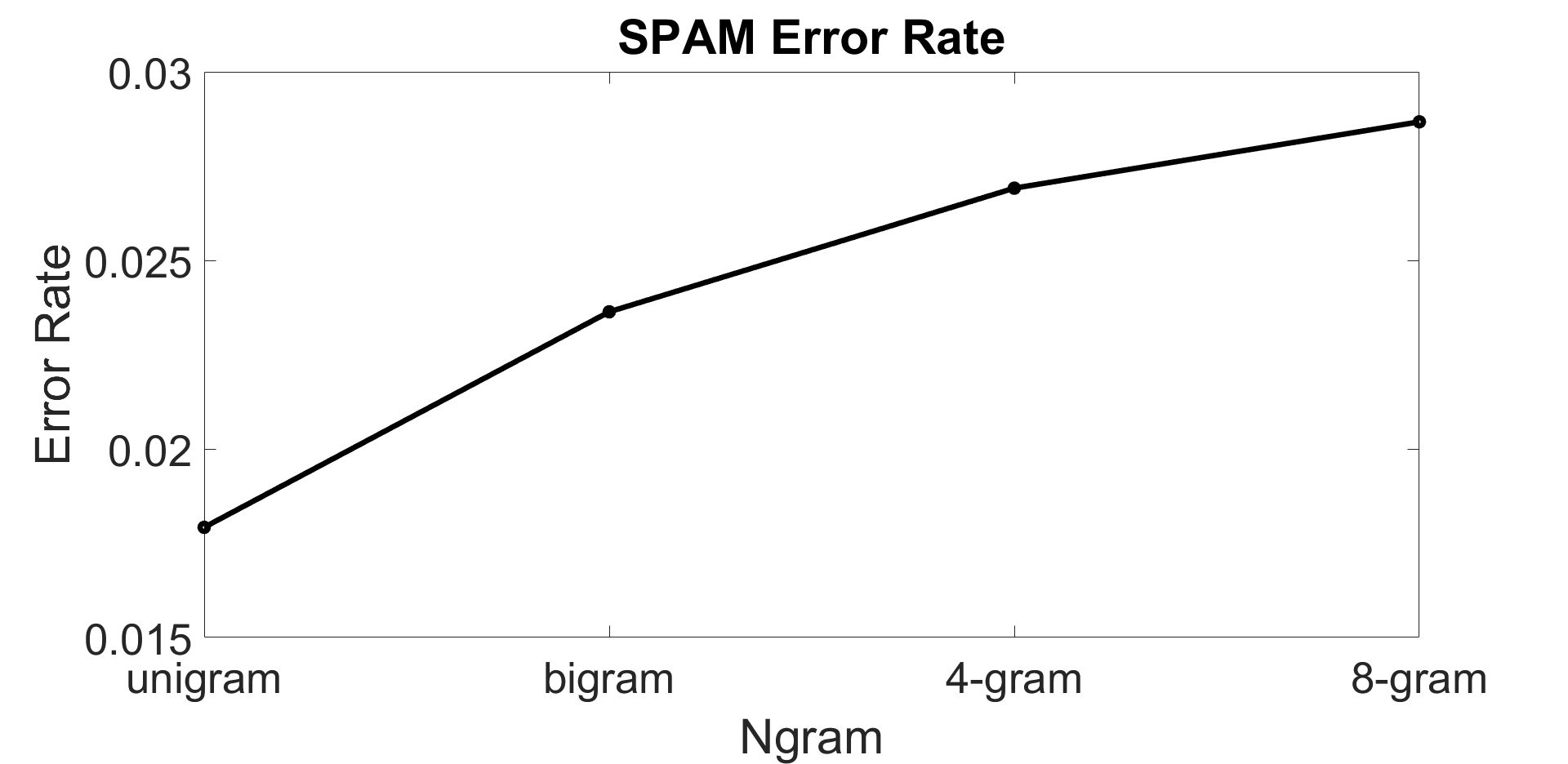}}
\caption{Comparison of error rates for the TMPCA method with N-gram 
preprocessed data where $N=1,2,4,8$.} \label{fig:ngrams}
\end{figure*}

The error rate results are shown in Fig. \ref{fig:ngrams}. We see that
the TMPCA method is robust with respect to the test dataset although it
was trained on the N-gram preprocessed data. This can be explained as
follows.  The TMPCA only examines the local property of grams in the
first tree-level.  In the following levels, the redundant information
between grams are removed, making the final reduced output contains only
semantic patterns. These patterns are also present in sentences which
are not preprocessed by the N-gram approach. The TMPCA performs the best
with the unigram in the SPAM error rate, with the bigram in the IMBD and
SST datasets, with the 4-gram in the SemEval dataset.

Overall, the TMPCA+SVM method achieves the state-of-the-art performance
that is commensurable or better than the RNN method. Such a dimension
reduction technique is very attractive by considering its less computing
power, shorter processing time and lower error rates.  The same
technique could be beneficial to other NLP problems as well. 

\section{Conclusion}\label{sec:conclusion}

A novel data processing technique called the TMPCA was proposed for text
classification problems in this work. The TMPCA can efficiently reduce the
dimension of the entire sentence data to facilitate the machine learning
task that follows. The complexity of the TMPCA was analyzed mathematically
to demonstrate its computational efficiency over the traditional PCA.
Furthermore, the classifiers need less training time to fit the TMPCA
processed data due to dimension reduction.  Finally, the TMPCA method
achieves the lowest error rates in all four datasets among benchmarking
methods. We would like to apply the TMPCA technique to other
challenging tasks and datasets as an extension of our current research efforts. 
 
\bibliographystyle{IEEEtran}
\bibliography{IEEEabrv}

\begin{thebibliography}{10}
\providecommand{\url}[1]{#1}
\csname url@samestyle\endcsname
\providecommand{\newblock}{\relax}
\providecommand{\bibinfo}[2]{#2}
\providecommand{\BIBentrySTDinterwordspacing}{\spaceskip=0pt\relax}
\providecommand{\BIBentryALTinterwordstretchfactor}{4}
\providecommand{\BIBentryALTinterwordspacing}{\spaceskip=\fontdimen2\font plus
\BIBentryALTinterwordstretchfactor\fontdimen3\font minus
  \fontdimen4\font\relax}
\providecommand{\BIBforeignlanguage}[2]{{%
\expandafter\ifx\csname l@#1\endcsname\relax
\typeout{** WARNING: IEEEtran.bst: No hyphenation pattern has been}%
\typeout{** loaded for the language `#1'. Using the pattern for}%
\typeout{** the default language instead.}%
\else
\language=\csname l@#1\endcsname
\fi
#2}}
\providecommand{\BIBdecl}{\relax}
\BIBdecl

\bibitem{NB}
N.~Friedman, G.~Dan, and G.~Moises, ``Bayesian network classifiers,''
  \emph{Machine learning}, vol.~29, pp. 131--163, 1997.

\bibitem{Sparsity}
Y.~Bengio, R.~Ducharme, P.~Vincent, and C.~Jauvin, ``A neural probabilistic
  language model,'' \emph{Journal of Machine Learning Research}, pp.
  1137--1155, 2003.

\bibitem{Conv_text}
X.~Zhang, Z.~Junbo, and Y.~LeCun, ``Character-level convolutional networks for
  text classification,'' \emph{Advances in neural information processing
  systems}, pp. 649--657, 2015.

\bibitem{LSTM}
S.~Hochreiter and J.~Schmidhuber, ``Long short-term memory,'' \emph{Neural
  Computation}, vol.~9, pp. 1735--1780, 1997.

\bibitem{GRU}
K.~Cho, B.~v. Merrienboer, C.~Gulcehre, D.~Bahdanau, F.~Bougares, H.~Schwenk,
  and Y.~Bengio, ``Learning phrase representations using rnn encoder¨cdecoder
  for statistical machine translation,'' \emph{Proc. {EMNLP}'2014}, 2014.

\bibitem{Time}
J.~Elman, ``Finding structure in time,'' \emph{Cognitive Science}, vol.~14, pp.
  179--211, 1990.

\bibitem{DBRNN-ELSTM}
\BIBentryALTinterwordspacing
Anonymous, ``Dependent bidirectional rnn with super-long short-term memory,''
  \emph{unpublished}, 2018. [Online]. Available:
  \url{https://openreview.net/forum?id=r1AMITFaW}
\BIBentrySTDinterwordspacing

\bibitem{Seq2Seq}
I.~Sutskever, O.~Vinyals, and Q.~V. Le, ``Sequence to sequence learning with
  neural networks,'' \emph{Advances in Neural Information Processing Systems},
  pp. 3104--3112, 2014.

\bibitem{Grammar}
O.~Vinyals, L.~Kaiser, T.~Koo, S.~Petrov, I.~Sutskever, and G.~Hinton,
  ``Grammar as a foreign language,'' \emph{Advances in Neural Information
  Processing Systems}, pp. 2773--2781, 2015.

\bibitem{LSA}
S.~Deerwester, T.~S. Dumais, W.~G. Furnas, K.~T. Landauer, and R.~Harshman,
  ``Indexing by latent semantic analysis,'' \emph{Journal of the American
  society for information science}, vol.~41, p. 391, 1990.

\bibitem{Noise_reduction}
Y.~Yang, ``Noise reduction in a statistical approach to text categorization,''
  in \emph{Proceedings of the 18th annual international ACM SIGIR conference on
  Research and development in information retrieval}.\hskip 1em plus 0.5em
  minus 0.4em\relax ACM, 1995, pp. 256--263.

\bibitem{Two-stage}
H.~Uguz, ``A two-stage feature selection method for text categorization by
  using information gain, principal component analysis and genetic algorithm,''
  \emph{Knowledge-Based Systems}, vol.~24, pp. 1024--1032, 2011.

\bibitem{Clustering}
L.~D. Baker and A.~K. McCallum, ``Distributional clustering of words for text
  classification,'' in \emph{Proceedings of the 21st annual international ACM
  SIGIR conference on Research and development in information retrieval}.\hskip
  1em plus 0.5em minus 0.4em\relax ACM, 1998, pp. 96--103.

\bibitem{SST_NLP_tool}
Standford, ``Corenlp,'' \url{https://stanfordnlp.github.io/CoreNLP/}.

\bibitem{Wiki2vec}
Google, ``Wiki2vec,'' \url{https://github.com/idio/wiki2vec}.

\bibitem{AdaGrad}
Duchi, ``Adaptive subgradient methods for online learning and stochastic
  optimization,'' \emph{The Journal of Machine Learning Research}, pp.
  2121--2159, 2011.

\bibitem{Ngrams}
B.~Peter~F., P.~V. Desouza, R.~L. Mercer, V.~J.~D. Pietra, and J.~C. Lai,
  ``Class-based n-gram models of natural language,'' \emph{Computational
  linguistics}, vol.~18, pp. 467--479, 1992.

\end{thebibliography}

\end{document}